%% file: arxiv.tex
\documentclass{article}
\PassOptionsToPackage{numbers, compress}{natbib}
\usepackage[preprint]{neurips_2026}

\usepackage[utf8]{inputenc}
\usepackage[T1]{fontenc}
\usepackage{hyperref}
\usepackage{url}
\usepackage{booktabs}
\usepackage{amsfonts}
\usepackage{nicefrac}
\usepackage{microtype}
\usepackage{xcolor}
\usepackage{amsmath,amssymb,amsthm}
\usepackage{graphicx}
\usepackage{multirow}
\usepackage{enumitem}
\usepackage{mathtools}
\usepackage{subcaption}
\usepackage{caption}

\definecolor{annblue}{RGB}{45,175,190}
\definecolor{annred}{RGB}{235,105,75}
\usepackage{xcolor}
\newcommand{\upblue}[1]{\hspace{0.5pt}{\raisebox{-0.35ex}{\scriptsize\textcolor{annblue}{$\uparrow$#1}}}}
\newcommand{\downred}[1]{\hspace{0.5pt}{\raisebox{-0.35ex}{\scriptsize\textcolor{annred}{$\downarrow$#1}}}}

\title{TCP-MCP: Landscape-Guided Co-Evolution of Prompts and Communication Topologies for Multi-Agent Systems}

\author{%
\textbf{Yi Ding$^{1}$ \quad
Zijie Xuan$^{2}$ \quad
Haowei Zhou$^{3}$ \quad
Zhenyu Ju$^{4}$ \quad
Xiaoxiao Dong$^{1}$} \\
\textbf{Jingwen Zhang$^{5}$ \quad
Xingyu Zhu$^{6}$\thanks{Corresponding author.} \quad
Leixin Sun$^{7}$ \quad
Haochi Zhang$^{8}$} \\
$^{1}$National Institute of Metrology, China \\
$^{2}$University of California, Berkeley \\
$^{3}$Shenzhen Institutes of Advanced Technology, Chinese Academy of Sciences \\
$^{4}$Nanjing University of Chinese Medicine \\
$^{5}$WEEX Exchange \\
$^{6}$National University of Singapore \\
$^{7}$Wuhan University \\
$^{8}$Peking University \\
\texttt{dingyiapply@163.com}
}

\begin{document}

\maketitle

\begin{abstract}
Effective multi-agent systems cannot be designed by selecting prompts or communication graphs in isolation. Agent behavior depends on the information an agent receives, while the usefulness of a communication edge depends on how the receiving agent interprets and uses that information. We propose \textbf{TCP-MCP} (Topology-Coupled Prompting for Multi-Agent Collaborative Problem-Solving), a co-evolution framework that searches agent prompts and communication topologies as a unified genome. TCP-MCP uses an initialization-time landscape probe to calibrate early search behavior, and then relies on Pareto-front diagnostics to adapt exploration under three objectives: task performance, token cost, and structural complexity. Using the same DeepSeek-V3.2 backbone across all methods, TCP-MCP achieves 82.66\%, 89.96\%, and 96.61\% accuracy on MMLU-Pro, MMLU, and GSM8K, respectively. Across the three benchmarks, it consistently outperforms automated graph-generation baselines and achieves competitive accuracy relative to debate-style systems, while using up to 5.69$\times$ fewer tokens than those systems at the reported operating points. These results show that jointly evolving prompts and communication structure provides a practical route to cost-aware and task-adaptive multi-agent system design in controlled evaluations.
\end{abstract}

\section{Introduction}

LLM-based multi-agent systems (MAS) support reasoning by assigning roles, exchanging intermediate reasoning, and aggregating outputs. As these systems become more capable, performance depends not only on each agent's prompt, but also on how agents are organized, what information passes between them, and which topology governs exchange. MAS construction is therefore a system-level design problem, rather than a direct extension of single-agent prompt engineering.

Current MAS design workflows remain fragmented. Single-agent reasoning methods such as Complex CoT~\cite{fu2023complexitybased} and Self-Consistency~\cite{wang2023selfconsistency} improve inference within one model. Multi-agent protocols such as LLM-Debate~\cite{du2024multiagentdebate} and LLM-Blender~\cite{jiang2023llmblender} introduce interaction and aggregation. General frameworks such as AutoGen~\cite{wu2024autogen}, ChatDev~\cite{qian2024chatdev}, and MetaGPT~\cite{hong2024metagpt} simplify workflow construction, while topology-oriented methods such as GPTSwarm~\cite{zhuge2024gptswarm} and G-Designer~\cite{zhang2025gdesigner} automate part of the communication-structure design process. These approaches, however, often fix the communication structure, optimize topology separately from prompts, or rely on reusable templates. This separation misses a central property of MAS design: prompt behavior and topology are coupled. A prompt that works under one communication pattern may fail under another, and an edge is useful only when the receiving agent is instructed to interpret and use the transmitted reasoning. We provide a more detailed discussion of related work in Appendix~\ref{app:relatedworks}. This coupling raises two challenges: how to move beyond fixed architectural priors and optimize communication structures together with prompts, and how to search the high-dimensional prompt--topology space without converging too early.

We propose TCP-MCP, a landscape-guided co-evolution framework that treats prompts and communication topology as a coupled object. TCP-MCP defines topology edits, prompt inheritance, mutation, and runtime prompt instantiation over a unified prompt--topology genome. It returns a Pareto set of MAS designs that trade off task performance, token cost, and structural complexity.

\paragraph{Contributions.}
Our contributions are as follows:
\begin{itemize}
    \item \textbf{Problem Identification.}
    We formulate joint prompt and topology optimization for multi-agent systems as a simultaneous co-evolution problem over a unified genome. This view treats MAS design as a coupled optimization problem over agent prompts and communication topology, and explains why isolated prompt or topology search can miss interactions induced by information flow. The resulting search space is inherently hybrid and task dependent, combining discrete structural decisions with semantic prompt choices.

    \item \textbf{\textit{Practical Solution.}} We propose TCP-MCP, a landscape-informed co-evolution framework that treats prompts and communication topology as a coupled design object. TCP-MCP uses an initialization-time landscape probe to calibrate early search behavior, and then adapts exploration using Pareto-front diagnostics computed across generations. The framework combines coupled evolutionary operators, Pareto-diagnostic adaptive control, and multi-objective selection over task performance, token cost, and structural complexity.

    \item \textbf{\textit{Experimental Validation.}} We evaluate TCP-MCP on MMLU, MMLU-Pro, and GSM8K under a unified backbone, showing that it outperforms strong baselines such as G-Designer. The results further show that TCP-MCP discovers task-adaptive collaboration structures while exposing clear Pareto trade-offs among performance, cost, and complexity.
\end{itemize}
\section{Preliminaries}

\subsection{Notations and Problem Formulation}

We model a multi-agent system (MAS) as a directed graph $G=(V,E)$, where each node is a role-specific agent and each edge specifies a communication link. Each node $v_i \in V$ is associated with a role--prompt pair $(role_i, p_i)$, and a candidate system is represented by a unified genome:
\begin{equation}
M = (G, P),
\end{equation}
where $P$ contains node-level role--prompt assignments. In TCP-MCP, prompt templates are stored independently of topology, but are instantiated at runtime with topology-dependent context:
\begin{equation}
\tilde{p}_i = \mathrm{BuildPrompt}\left(p_i, N^{-}(v_i;G)\right),
\end{equation}
where $N^{-}(v_i;G)$ is the incoming neighborhood of $v_i$. As a result, the behavior of a node can change with $G$ even when its stored template $p_i$ is inherited without modification. The final prediction is produced by a decision node $v_{dec}$, which aggregates the relevant outputs from sink nodes.

Prompt behavior depends on communication context, and the utility of an edge depends on the behavior induced at the receiving node. MAS design is therefore a coupled optimization problem over prompts and topology. Given an evaluation set $D$, we optimize task accuracy, token cost, and structural complexity. Let $A(M)$ denote task accuracy, $C(M)$ denote token cost, and $K(G)=|V|+|E|$ denote graph complexity. We define the multi-objective maximization problem as:
\begin{equation}
F(M)=\left(A(M), -C(M), -\ln(1+K(G))\right).
\label{eq:fitness_vector}
\end{equation}
The logarithmic penalty reduces the effect of large graph-size differences and prevents structural complexity from dominating the Pareto ranking during multi-objective environmental selection.

\subsection{A Coarse Landscape Signal}

The unified genome defines a hybrid search space $\mathcal{S}=\mathcal{G}\times\Theta$, where $\mathcal{G}$ is the set of feasible topologies and $\Theta$ is the space of role--prompt configurations. This space is combinatorial, high-dimensional, rugged, and multimodal, with coupled structural and semantic variables. TCP-MCP therefore uses a coarse landscape signal during initialization to calibrate the early search behavior.

We compute a scalar preference score $S(M)$ that rewards accuracy while penalizing token cost and normalized graph complexity during the initial landscape probe:
\begin{equation}
S(M)=\frac{A(M)^k}{(C(M)/T_0)^\gamma \cdot \bar{K}(G)^\beta},
\end{equation}
where $T_0$ normalizes cost, $\bar{K}(G)=1+K(G)/K_0$ is a graph complexity term, $K_0>0$ normalizes graph complexity, and $k,\gamma,\beta>0$ control the relative emphasis of the objectives. Fitness Distance Correlation (FDC) is then computed over the initialized population $\mathcal{P}_0=\{M_j\}_{j=1}^{n_0}$ as:
\begin{equation}
\mathrm{FDC}=\mathrm{Spearman}\left(
\{S(M_j)\}_{j=1}^{n_0},
\{d(M_j,M^\star)\}_{j=1}^{n_0}
\right),
\end{equation}
where $M^\star=\arg\max_{M_j\in\mathcal{P}_0}S(M_j)$ is the current best individual under $S(M)$ and $d(M_j,M^\star)$ denotes genome distance to $M^\star$. We use FDC only for initialization-time landscape probing; a lower or negative correlation means that higher-preference candidates are closer to the current best individual. After initialization, TCP-MCP relies on cross-generational Pareto-front diagnostics for adaptation instead of recomputing this coarse landscape signal.

\begin{figure}[!t]
    \centering
    \includegraphics[width=0.99\linewidth]{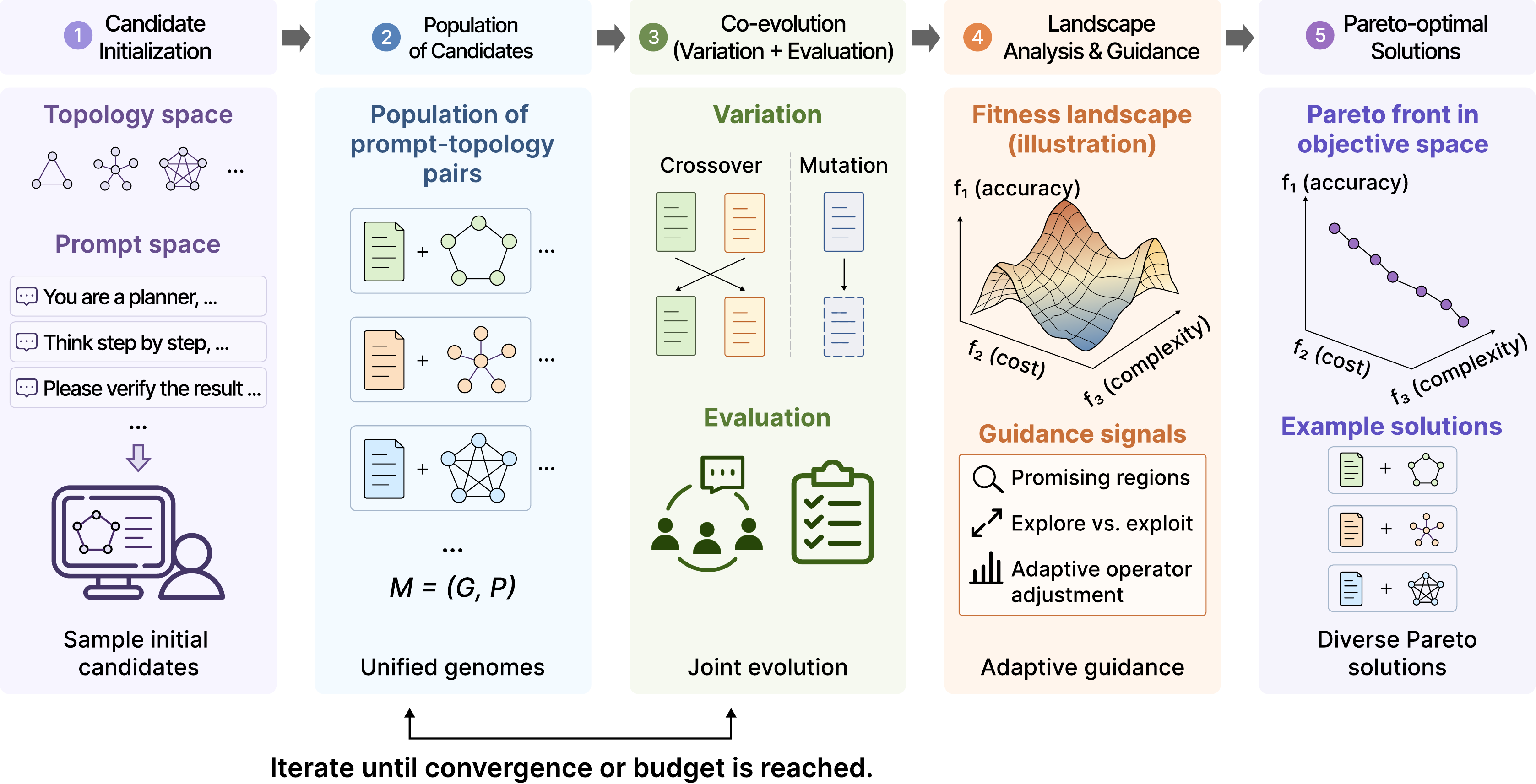}
    \caption{Overall framework of TCP-MCP for joint prompt--topology co-evolution. Each candidate is represented as a unified prompt--topology genome. The numbered blocks show the main functional modules; the text groups them into initialization, co-evolution, and final Pareto selection.}
    \label{fig:framework}
\end{figure}
\section{Methods}

\subsection{Overall Framework}

As shown in Figure~\ref{fig:framework}, TCP-MCP formulates MAS construction as a system-level design problem. Each candidate is encoded as a unified genome $M=(G,P)$, where $G$ specifies the communication topology and $P$ contains node-level role and prompt assignments. This representation keeps topology and prompting coupled rather than optimizing them in isolation. 
The framework proceeds in three stages. It first builds a diverse population with Latin Hypercube Sampling (LHS)~\cite{mckay1979lhs}, instantiates the sampled prompt--topology configurations as executable systems, and computes FDC~\cite{jones1995fdc} as a coarse initialization signal. Details of the initial topology and prompt construction are given in Appendix~\ref{app:initial_population}. TCP-MCP then co-evolves prompts and topology. In each generation, candidates are evaluated with a multi-objective fitness vector, selected by elitist environmental selection~\cite{deb2002nsga2}, and updated through structural crossover and mutation. Pareto-front diagnostics~\cite{zitzler2000comparison,deb2001multiobjective} are monitored throughout evolution to guide exploration and preserve coverage across regions. The final output is a Pareto elite set over task performance, token cost, and structural complexity, from which we select a representative development-set operating point for held-out evaluation.

\subsection{Structural Crossover and Prompt Handling Mechanism}

TCP-MCP uses Prompt Minimal Inheritance (PMI): crossover transfers topological subgraphs rather than prompt fragments. Recipient-parent nodes keep their prompt templates, while transplanted nodes inherit donor templates when available and otherwise use lightweight defaults.

Given two parent genomes, $M_a=(G_a,P_a)$ and $M_b=(G_b,P_b)$, TCP-MCP selects anchor nodes to define crossover modules. It first searches for exact matches between canonical role names. If none exist, it uses the boundary-degree heuristic:
\begin{equation}
b(v)=\mathrm{in\_degree}(v)+\mathrm{out\_degree}(v),
\end{equation}
and samples high-ranked boundary nodes as module interfaces. The selected subgraphs are then exchanged and stitched into the recipient topology.

Prompt construction is deferred to runtime. Although crossover changes the communication graph, stored prompt templates are not rewritten. Instead, topology-dependent context is added during execution through $\mathrm{BuildPrompt}(\cdot)$, keeping templates stable while allowing node behavior to adapt to the current graph and its incoming messages after structural changes.

\begin{figure}[!t]
    \centering
    \includegraphics[width=0.99\linewidth]{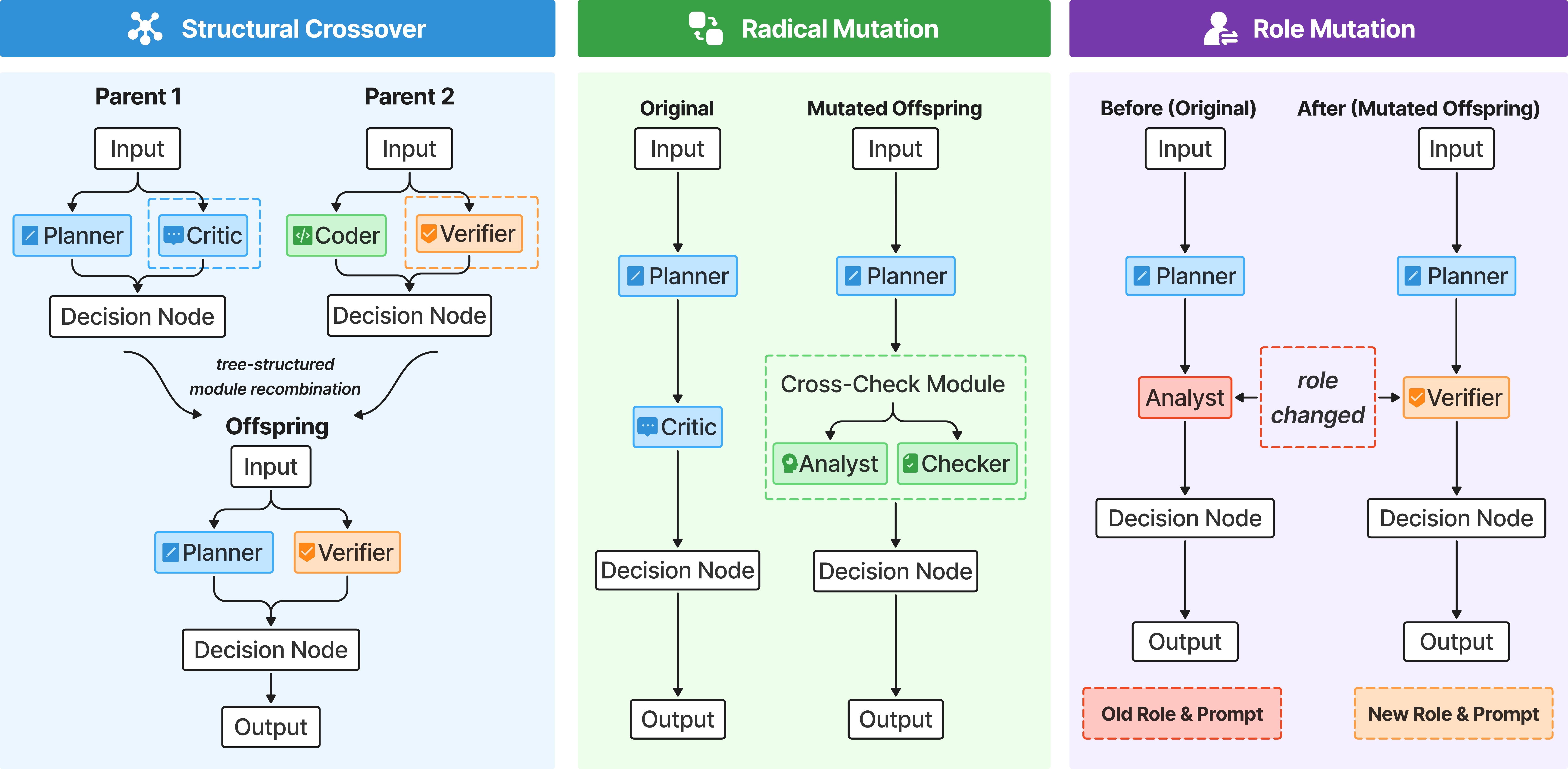}
    \caption{Illustration of structural crossover, radical mutation, and role mutation in TCP-MCP. The structural crossover panel uses PMI. }
    \label{fig:crossover}
\end{figure}

\subsection{Mutation Operators}

After structural crossover, TCP-MCP mutates the unified genome to maintain exploration beyond module recombination. Given a genome $M=(G,P)$, mutation produces an offspring:
\begin{equation}
M' = \mathcal{M}(M),
\end{equation}
through a probabilistic scheme. Topology mutation and role mutation are sampled as stochastic operators, while radical mutation is activated with a probability determined by stagnation signals.

\textbf{Topology mutation.} Topology mutation applies local edits to the communication graph,
\begin{equation}
\mathcal{M}_{\mathrm{topo}} \in 
\{\textsc{AddEdge},\ \textsc{RemoveEdge},\ \textsc{AddNode},\ \textsc{RemoveNode}\},
\end{equation}
with sampling weights adjusted during search. Edits that create cycles are rejected and retried. If repeated attempts fail, the genome is left unchanged.

\textbf{Role mutation.} Role mutation changes identities using a three-tier role pool. When a role changes, its prompt is regenerated by \texttt{PromptMutator} so that the new role and prompt remain aligned.

\textbf{Radical mutation.} Radical mutation introduces larger structural changes when the search appears stagnant. Its activation is controlled by stagnation signals computed from Pareto-front diagnostics, rather than by a fixed mutation schedule. The corresponding activation probability is defined in Section~\ref{main:Adaptivecontrol}, and further implementation details are given in Appendix~\ref{app:mutation_details}.
\subsection{Environmental Selection and Diversity Preservation}
\label{sec:environmental_selection}

At each generation, TCP-MCP applies elitist environmental selection after offspring generation. Selection follows NSGA-II with constraint dominance, together with additional mechanisms for preserving diversity across cost and structural regimes in the elite population.
Let the fitness vector be defined as in Eq.~\ref{eq:fitness_vector}. Feasibility is determined by a conservative dynamic accuracy floor:
\begin{equation}
\tau=\mathrm{best\_acc}(t)-\delta,
\end{equation}
where $\mathrm{best\_acc}(t)$ is the best accuracy observed at generation $t$ and $\delta$ is a fixed tolerance. This threshold reduces feasibility oscillation when the best individual changes. Feasible candidates are prioritized under constrained dominance, with tie-breaking details given in Appendix~\ref{app:selection_details}.

After non-dominated sorting, TCP-MCP preserves diversity in the cost and structure space through bin-based niching. Candidates are assigned to quantile-based bins over token cost and structural complexity, and each bin has a capped occupancy to avoid crowding around similar trade-off points. Any remaining elite slots are filled according to the original NSGA-II ranking.

TCP-MCP also imposes a structural quota on the elite set to avoid collapse toward trivial low-cost solutions. The quota retains complex candidates if they satisfy the accuracy floor. Without this constraint, preliminary runs can collapse to 1--2 node topologies that exploit the cost objective and weaken the Pareto front. An external elite archive improves diversity and robustness by periodically reintroducing high-performing genomes. Detailed settings are provided in Appendix~\ref{app:selection_details}.

Together, these mechanisms maintain a stable and diverse Pareto set across performance, token cost, and structural complexity throughout evolutionary search under changing conditions.

\subsection{Landscape Perception and Adaptive Control}
\label{main:Adaptivecontrol}
TCP-MCP separates landscape perception into initialization-time calibration and evolution-time response. At initialization, FDC provides a coarse signal for calibrating the early crossover bias. During evolution, TCP-MCP does not re-estimate FDC; it instead monitors cross-generational Pareto diagnostics, including normalized HV, spacing, frontier gaps, and regional coverage. These diagnostics are combined into a response index that drives adaptive exploration and injection decisions:
\begin{equation}
s = w_{\mathrm{HV}}s_{\mathrm{HV}} + w_{\mathrm{acc}}s_{\mathrm{acc}} + w_{\mathrm{tok}}s_{\mathrm{tok}} + w_{\mathrm{div}}s_{\mathrm{div}},
\end{equation}
which summarizes hypervolume stagnation, accuracy stagnation, cost pressure, and diversity deficit. The response index is mapped to an exploration probability:
\begin{equation}
p_{\mathrm{stag}} = \sigma(\alpha s + \beta),
\label{eq:stagnation_probability}
\end{equation}
which increases mutation and injection pressure when the frontier stagnates or diversity declines. Detailed normalization and response rules are given in Appendix~\ref{app:adaptive_control}.

\section{Experiments}
\label{sec:experiments}

We evaluate TCP-MCP along four axes: Q1 (Superiority), Q2 (Resilience), Q3 (Effectiveness), and Q4 (Sensitivity). Q1--Q3 are reported in this section, and additional discussion of sensitivity and design choices, including implementation-constant and budget effects, is provided in the appendix.

\subsection{Experimental Setup}

\paragraph{Datasets and Evaluation.}
We evaluate TCP-MCP on MMLU~\citep{hendrycks2021mmlu}, MMLU-Pro~\citep{wang2024mmlupro}, and GSM8K~\citep{cobbe2021training}, covering broad-domain question answering, high-distraction option analysis, and multi-step mathematical reasoning. For search, we sample fixed development subsets of 80, 100, and 40 questions for MMLU, MMLU-Pro, and GSM8K from the original test collections. The final evaluation set is each original test collection minus its development subset, making development and test sets strictly disjoint. Final results are reported only on this held-out set. We use accuracy and token cost as main metrics, and track normalized Hypervolume (HV) and frontier trends to analyze search. Auxiliary scalar signals, such as the preference score, are used only for landscape probing.

\paragraph{Baselines and Fairness Control.}
We compare TCP-MCP with DeepSeek-V3.2 Zero-shot, Complex CoT~\cite{fu2023complexitybased}, Self-Consistency~\cite{wang2023selfconsistency}, and PHP~\cite{zheng2024progressivehint} as single-agent or single-workflow reasoning baselines; Complete Graph and Random Graph as fixed or sampled topology MAS baselines; LLM-Debate~\cite{du2024multiagentdebate} and LLM-Blender~\cite{jiang2023llmblender} as multi-agent aggregation or debate baselines; AutoGen~\cite{wu2024autogen}, ChatDev~\cite{qian2024chatdev}, and MetaGPT~\cite{hong2024metagpt} as general multi-agent workflow frameworks; and GPTSwarm~\cite{zhuge2024gptswarm} and G-Designer~\cite{zhang2025gdesigner} as automated topology-design baselines. For the main comparisons in Table~\ref{tab:main_results}, all methods use the same backbone, DeepSeek-V3.2, with the same API endpoint, decoding configuration, and context budget. Unless otherwise specified, we use the official DeepSeek API defaults without manually overriding decoding hyperparameters: temperature $=1.0$, top-p $=1.0$, frequency penalty $=0$, and presence penalty $=0$. For baselines involving structural search, we rerun them under the same evaluation protocol and comparable budget constraints.

\paragraph{Search and Reporting Protocol.}
On each dataset, TCP-MCP starts from an initial population of 16 candidate systems and performs multi-objective search across generations. Environmental selection retains 8 elites per generation. Unless stated, all experiments use the same population and elite settings. We report the generation-21 Pareto operating point, where HV trends indicate a stabilized Pareto front. Since TCP-MCP returns a Pareto set rather than a single accuracy-only system, we select the reported solution using a fixed development-set rule that filters non-dominated candidates by an accuracy floor, removes high-cost tail points without meaningful development-accuracy gains, and breaks ties by token cost and complexity before held-out evaluation. Details of the rule are provided in Appendix~\ref{app:operating_point_selection}. Table~\ref{tab:main_results} reports held-out inference tokens; Appendix~\ref{app:compute_resources} reports held-out API-call counts, search-time tokens, search-time API calls, and wall-clock cost.

\begin{table*}[t]
\centering
\scriptsize
\caption{Main results on MMLU, MMLU-Pro, and GSM8K. Accuracy is reported in percentage points, and token usage is reported in millions. Small annotations indicate baseline-minus-TCP-MCP differences; blue upward arrows denote larger values than TCP-MCP, and orange downward arrows denote smaller values.}
\label{tab:main_results}

\begin{tabular}{lcccccc}
\toprule
\multirow{2}{*}{\textbf{Method}}
& \multicolumn{2}{c|}{\textbf{MMLU}}
& \multicolumn{2}{c|}{\textbf{MMLU-Pro}}
& \multicolumn{2}{c}{\textbf{GSM8K}} \\
\cmidrule(lr){2-3}\cmidrule(lr){4-5}\cmidrule(lr){6-7}
& \textbf{Acc.} & \textbf{Tok.}
& \textbf{Acc.} & \textbf{Tok.}
& \textbf{Acc.} & \textbf{Tok.} \\
\midrule

Zero-shot
& 84.10\downred{5.86} & 3.42M\downred{84.61}
& 54.36\downred{28.30} & 2.79M\downred{94.49}
& 95.78\downred{0.83} & 0.21M\downred{5.81} \\

PHP
& 87.96\downred{2.00} & 9.13M\downred{78.90}
& 74.71\downred{7.95} & 14.46M\downred{82.82}
& 96.21\downred{0.40} & 0.71M\downred{5.31} \\

Complete Graph
& 85.92\downred{4.04} & 82.71M\downred{5.32}
& 73.95\downred{8.71} & 132.45M\upblue{35.17}
& 92.11\downred{4.50} & 7.65M\upblue{1.63} \\

GPTSwarm
& 81.70\downred{8.26} & 12.60M\downred{75.43}
& 74.23\downred{8.43} & 20.62M\downred{76.66}
& 77.71\downred{18.90} & 1.21M\downred{4.81} \\

LLM-Blender
& 87.23\downred{2.73} & 4.68M\downred{83.35}
& 63.55\downred{19.11} & 6.43M\downred{90.85}
& 92.87\downred{3.74} & 0.33M\downred{5.69} \\

G-Designer
& 87.90\downred{2.06} & 28.04M\downred{59.99}
& 79.32\downred{3.34} & 32.87M\downred{64.41}
& 95.81\downred{0.80} & 4.44M\downred{1.58} \\

Complex CoT
& 87.49\downred{2.47} & 4.87M\downred{83.16}
& 79.74\downred{2.92} & 11.04M\downred{86.24}
& 94.18\downred{2.43} & 0.45M\downred{5.57} \\

Self-Consistency
& 90.36\upblue{0.40} & 12.47M\downred{75.56}
& 76.12\downred{6.54} & 45.69M\downred{51.59}
& 96.05\downred{0.56} & 1.96M\downred{4.06} \\

LLM-Debate
& 90.84\upblue{0.88} & 404.09M\upblue{316.06}
& 83.67\upblue{1.01} & 553.41M\upblue{456.13}
& 95.48\downred{1.13} & 24.61M\upblue{18.59} \\

Random Graph
& 90.40\upblue{0.44} & 255.06M\upblue{167.03}
& 82.59\downred{0.07} & 331.48M\upblue{234.20}
& 95.81\downred{0.80} & 14.78M\upblue{8.76} \\

AutoGen
& 80.37\downred{9.59} & 2.27M\downred{85.76}
& 55.72\downred{26.94} & 3.18M\downred{94.10}
& 88.79\downred{7.82} & 0.39M\downred{5.63} \\

ChatDev
& 83.71\downred{6.25} & 4.14M\downred{83.89}
& 62.19\downred{20.47} & 5.47M\downred{91.81}
& 96.13\downred{0.48} & 0.44M\downred{5.58} \\

MetaGPT
& 86.20\downred{3.76} & 1.86M\downred{86.17}
& 59.03\downred{23.63} & 2.53M\downred{94.75}
& 93.06\downred{3.55} & 0.12M\downred{5.90} \\

\midrule
\textbf{TCP-MCP}
& \textbf{89.96} & \textbf{88.03M}
& \textbf{82.66} & \textbf{97.28M}
& \textbf{96.61} & \textbf{6.02M} \\

\bottomrule
\end{tabular}

\end{table*}
\subsection{Superiority (Q1)}
We evaluate selected Pareto solutions on held-out test sets against single-agent baselines, fixed-topology MAS variants, and automated design baselines under the same backbone and protocol. Table~\ref{tab:main_results} shows that TCP-MCP outperforms G-Designer, the strongest automated graph-generation baseline, on all three benchmarks, raising accuracy from 87.90 to 89.96 on MMLU, 79.32 to 82.66 on MMLU-Pro, and 95.81 to 96.61 on GSM8K. These gains use more tokens than G-Designer, so TCP-MCP is best viewed as an accuracy-oriented Pareto point rather than a cheaper replacement. Against high-cost protocols, TCP-MCP offers a favorable accuracy--cost trade-off: compared with LLM-Debate, it is 0.88 points lower on MMLU with 4.59$\times$ fewer tokens, 1.01 points lower on MMLU-Pro with 5.69$\times$ fewer tokens, and 1.13 points higher on GSM8K with 4.09$\times$ fewer tokens. On MMLU-Pro, it slightly exceeds Random Graph while using about 3.41$\times$ fewer tokens. Overall, TCP-MCP improves automated topology design and remains competitive with expensive multi-agent protocols, although its selected points are not always the lowest-cost solutions. The corresponding accuracy and token-cost trade-offs are visualized in Appendix Figure~\ref{fig:app_acc_token_tradeoff}.

\begin{figure*}[t]
    \centering
    \includegraphics[width=0.99\linewidth]{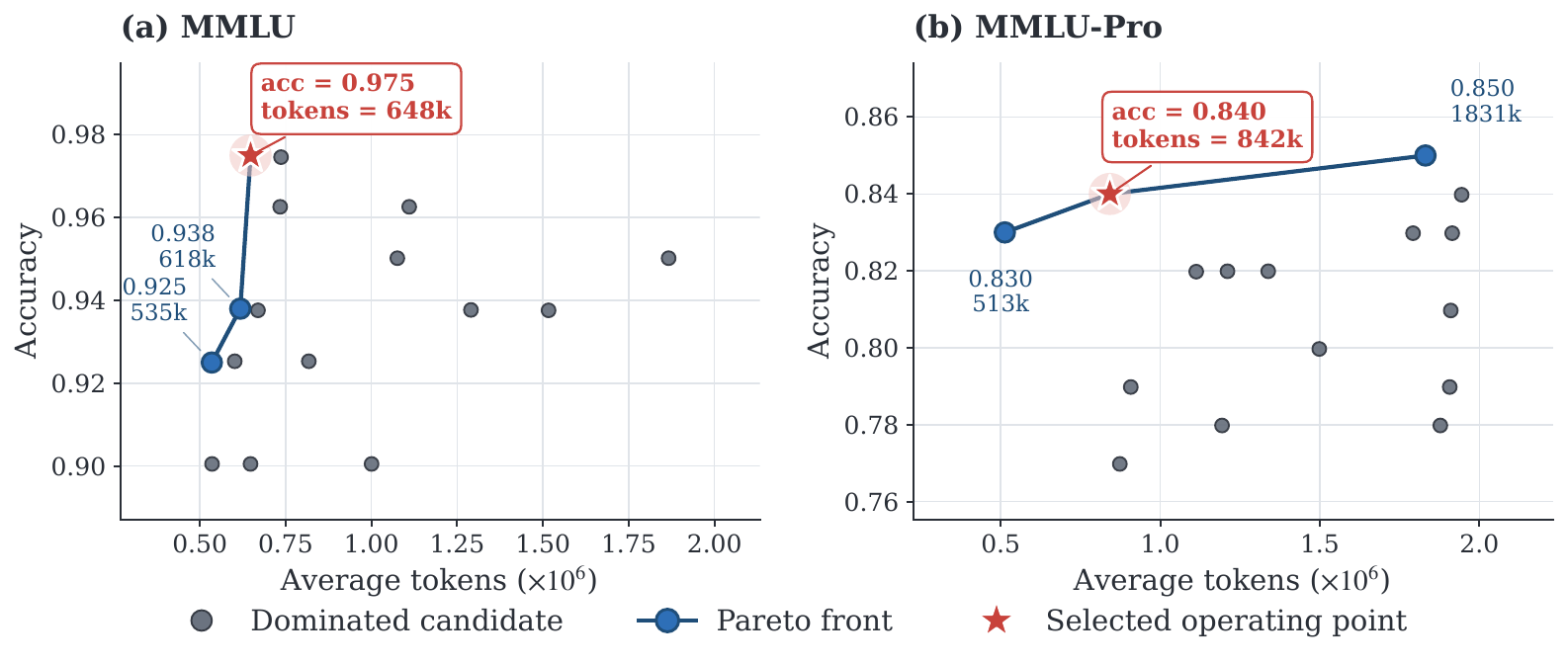}
    \caption{Development-set Pareto fronts. Stars mark the representative TCP-MCP points selected for held-out evaluation under the fixed development-set selection rule. The selected points are chosen from the development-set Pareto front by considering both accuracy and token cost. We show MMLU and MMLU-Pro because they contain visible trade-off fronts; GSM8K is omitted because the selected operating region does not form a visually informative trade-off front.}
    \label{fig:pareto_front}
\end{figure*}

Figure~\ref{fig:pareto_front} shows the development-set Pareto fronts used to choose the TCP-MCP operating points. The selection is made before held-out evaluation and uses development-set Pareto candidates. On MMLU, we select the highest-accuracy point with competitive token cost, while on MMLU-Pro we avoid a much higher-cost point that gives only a small accuracy gain. This reflects the multi-objective goal of TCP-MCP; the development-set selection rule is formalized in Appendix~\ref{app:operating_point_selection}.
\subsection{Resilience (Q2)}

We study TCP-MCP's resilience from two angles: Pareto-front stability during evolution and reuse of searched structures beyond their original search setting.

\begin{figure}[!t]
    \centering
    \includegraphics[width=0.99\linewidth]{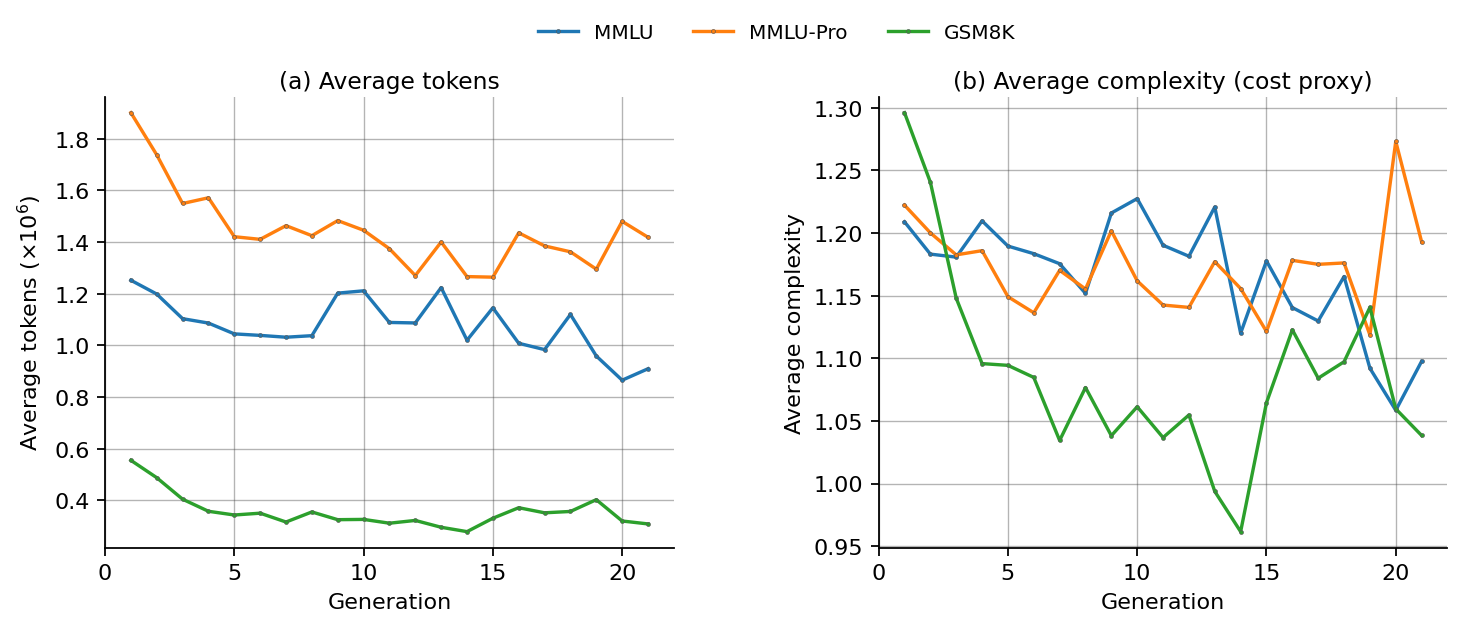}
    \caption{Evolutionary trends (Q2) showing token cost and structural complexity across generations.}
    \label{fig:trends}
\end{figure}

Figure~\ref{fig:trends} plots token cost and structural complexity across generations. Appendix~\ref{app:hv_diagnostics} reports normalized Hypervolume (HV) at generations 1, 11, and 21. These HV diagnostics show that the Pareto front improves mainly in early generations and stabilizes as evolution approaches generation 21. Average token usage generally decreases or stabilizes, while average structural complexity remains bounded rather than growing monotonically. Together, these diagnostics suggest that TCP-MCP does not improve the frontier simply by enlarging systems or spending more tokens.

We next examine whether the evolved structures remain useful outside their original search setting. Table~\ref{tab:cross_dataset_transfer} and Table~\ref{tab:gpt4_transfer} report two complementary transfer settings. In cross-dataset transfer between the two multiple-choice benchmarks, we apply a searched prompt--topology genome to another dataset without re-running search or using the target development set. The transferred genomes remain competitive, with accuracy drops of 1.54 points from MMLU to MMLU-Pro and 1.20 points from MMLU-Pro to MMLU. In cross-backbone transfer, we reuse topologies searched with DeepSeek-V3.2 under GPT-4. TCP-MCP achieves the best GPT-4 accuracy among the compared methods on all three benchmarks, with the largest gain on MMLU-Pro. Appendix~\ref{app:second_eval} also reports a second held-out inference run, where the selected systems follow the same overall trend as the main results.

\begin{table}[t]
\centering
\small
\renewcommand{\arraystretch}{1.05}

\begin{tabular}{@{}p{0.49\linewidth}@{\hspace{0.02\linewidth}}p{0.49\linewidth}@{}}

\begin{minipage}[t][6.2em][t]{\linewidth}
\centering
\captionof{table}{
Native rows are included as references; transfer rows evaluate a searched prompt--topology genome on a different target dataset without re-running evolutionary search.
}
\label{tab:cross_dataset_transfer}
\end{minipage}
&
\begin{minipage}[t][6.2em][t]{\linewidth}
\centering
\captionof{table}{
Cross-backbone transfer for resilience (Q2). For topology-based methods, topologies searched under DeepSeek-V3.2 are reused with GPT-4 without target-backbone search.
}
\label{tab:gpt4_transfer}
\end{minipage}
\\[-0.2em]

\centering
\setlength{\tabcolsep}{3.5pt}
\resizebox{\linewidth}{!}{
\begin{tabular}{llcc}
\toprule
\textbf{Source} & \textbf{Target} & \textbf{Acc.} & \textbf{Drop} \\
\midrule
MMLU & MMLU & 89.96 & -- \\
MMLU-Pro & MMLU-Pro & 82.66 & -- \\
MMLU & MMLU-Pro & 81.12 & 1.54 \\
MMLU-Pro & MMLU & 88.76 & 1.20 \\
\bottomrule
\end{tabular}
}
&
\centering
\setlength{\tabcolsep}{3.5pt}
\resizebox{\linewidth}{!}{
\begin{tabular}{lccc}
\toprule
\textbf{Method} & \textbf{MMLU} & \textbf{M-Pro} & \textbf{GSM8K} \\
\midrule
Vanilla & 82.14 & 55.35 & 85.40 \\
Self-Cons. & 83.65 & 69.88 & 86.14 \\
G-Designer & 84.50 & 74.79 & 86.14 \\
TCP-MCP & \textbf{85.48} & \textbf{78.63} & \textbf{96.53} \\
\bottomrule
\end{tabular}
}

\end{tabular}

\end{table}
Together, the trajectory, transfer, and second-evaluation analyses provide complementary evidence for resilience. The evolutionary curves show that, within TCP-MCP search, frontier improvement is not driven by monotonic growth in average token cost or structural complexity. The cross-dataset results indicate that searched genomes can retain strong performance on related benchmarks, while the GPT-4 transfer results suggest that some searched topologies preserve useful collaboration patterns across backbones. These findings support partial reuse across datasets and models, although target-native search remains preferable when development data and search budget are available. This resilience should not be interpreted as universal transferability. Instead, it indicates that the searched communication patterns capture reusable collaboration biases under related tasks and compatible backbones, while still leaving room for target-specific search when the deployment setting changes substantially, the task format shifts, or the objective priorities change.
\subsection{Effectiveness (Q3)}

We evaluate whether TCP-MCP learns task-suited collaboration structures and whether its search procedure contributes to selected solutions. We first examine role structures selected by the final Pareto elites, then analyze operator and coupling ablations. Operator ablations test TCP-MCP's internal search components, while coupling ablations test whether prompt--topology design is better optimized jointly than in sequential stages under the same evaluation and selection protocol.

The final Pareto elites show different structural patterns across tasks. On GSM8K, the selected system is compact: a mathematical analysis role interprets the problem before passing it to a solver. On MMLU, the structure emphasizes verification, combining knowledge checking with option-level validation before final solving. On MMLU-Pro, TCP-MCP anchors relevant definitions and decomposes the question before solving, which helps under stronger distractors. Although a solver role appears in all three cases, its prompt changes with the task. It acts as a verification coordinator on MMLU and MMLU-Pro, but remains closer to a direct solver on GSM8K.

\begin{figure}[!t]
    \centering
    \includegraphics[width=0.99\linewidth]{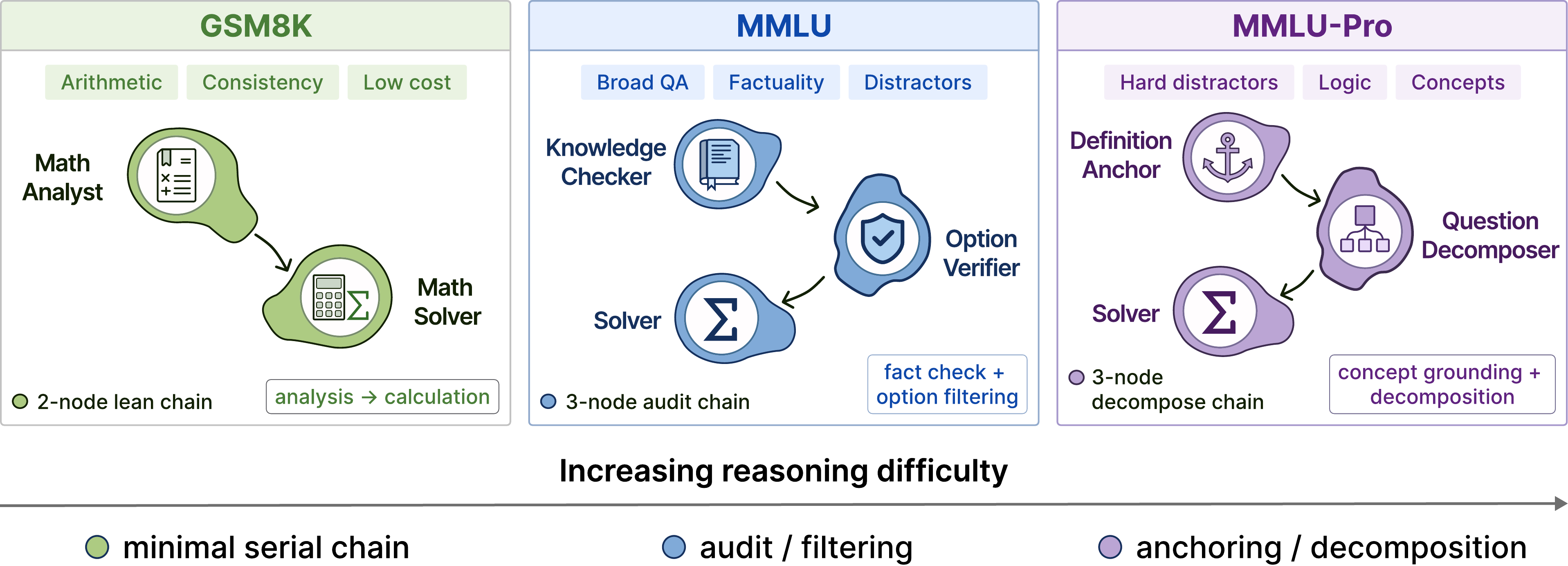}
    \caption{Task-adaptive structural prototypes evolved for different reasoning requirements.}
    \label{fig:prototypes}
\end{figure}

We ablate four variants of the genome search over $M=(G,P)$. AB1, AB2, and AB3 are operator ablations: AB1 removes adaptive control, AB2 removes crossover, and AB3 removes mutation-related exploration. AB4 is a staged coupling ablation: it first optimizes prompts under a fixed topology, then fixes the prompt configuration and searches topology. This variant tests whether prompt--topology design can be separated into sequential phases. Full TCP-MCP keeps topology updates, role assignment, prompt inheritance, and prompt mutation coupled throughout evolution.

As shown in Table~\ref{tab:ablation_all}, full TCP-MCP obtains the best selected accuracy on all three benchmarks. Among operator ablations, removing structural crossover yields the largest drop on MMLU-Pro (-2.51pp), suggesting that local perturbations do not replace module-level recombination. Removing mutation gives the largest drop on MMLU (-2.59pp) and uses 1.46$\times$ more tokens than the full model, so the weaker result cannot be attributed to a smaller inference budget. AB1 has mixed effects: its drop is small on MMLU-Pro and GSM8K but larger on MMLU, suggesting that adaptive control mainly stabilizes search while its benefit varies by dataset.

The staged variant AB4 underperforms full TCP-MCP across the three benchmarks. This suggests that a prompt-first, topology-second sequential procedure is weaker than simultaneous co-evolution in this setting, where each side constrains the other's useful updates. Together, these ablations support TCP-MCP's search components and the treatment of prompts and communication topology as an interdependent design object under the evaluated search and reporting protocol.

\begin{table*}[t]
\centering
\scriptsize
\caption{
Ablation analysis for effectiveness and the staged optimization variant (Q3).
Token usage follows Table~\ref{tab:main_results}: held-out inference tokens for the selected operating point, reported in millions.
}
\label{tab:ablation_all}

\begin{tabular*}{\textwidth}{@{}p{0.66\textwidth}@{\extracolsep{\fill}}p{0.32\textwidth}@{}}

\centering
\textbf{(a) Ablation Summary: Acc / $\Delta$ / Tok. (M)}
\vspace{0.3em}

\setlength{\tabcolsep}{2pt}
\renewcommand{\arraystretch}{1.05}
\begin{tabular*}{\linewidth}{@{\extracolsep{\fill}}lccc@{}}
\toprule
\textbf{Setting} & \textbf{MMLU} & \textbf{MMLU-Pro} & \textbf{GSM8K} \\
\midrule
Full & \textbf{89.96 / -- / 88.03} & \textbf{82.66 / -- / 97.28} & \textbf{96.61 / -- / 6.02} \\
AB1 & 88.23 / -1.73 / 89.43 & 82.48 / -0.18 / 74.46 & 95.97 / -0.64 / 6.14 \\
AB2 & 87.70 / -2.26 / 65.67 & 80.15 / \textbf{-2.51} / 230.59 & 95.65 / -0.97 / 5.65 \\
AB3 & 87.37 / \textbf{-2.59} / 128.55 & 81.46 / -1.20 / 222.17 & 96.29 / -0.32 / 6.51 \\
AB4 & 87.48 / -2.48 / 87.78 & 82.30 / -0.36 / 98.55 & 96.27 / -0.35 / 6.23 \\

\bottomrule
\end{tabular*}

&
\centering
\textbf{(b) Token Ratio (relative to Full)}
\vspace{0.3em}

\setlength{\tabcolsep}{2pt}
\renewcommand{\arraystretch}{1.26}
\begin{tabular*}{\linewidth}{@{\extracolsep{\fill}}lccc@{}}
\toprule
\textbf{Setting} & \textbf{MMLU} & \textbf{M-Pro} & \textbf{GSM8K} \\
\midrule
AB1 / Full & 1.02$\times$ & 0.77$\times$ & 1.02$\times$ \\
AB2 / Full & 0.75$\times$ & \textbf{2.37$\times$} & 0.94$\times$ \\
AB3 / Full & \textbf{1.46$\times$} & \textbf{2.28$\times$} & 1.08$\times$ \\
AB4 / Full & 1.00$\times$ & 1.01$\times$ & 1.03$\times$ \\
\bottomrule
\end{tabular*}

\end{tabular*}

\end{table*}
\section{Conclusion, Limitations, and Future Work}

\noindent{\textbf{Conclusion.}}
We presented TCP-MCP, a co-evolution framework that optimizes prompts and communication topologies as an interdependent genome under performance, cost, and complexity constraints. Results on MMLU, MMLU-Pro, and GSM8K suggest that MAS design is better treated as coupled optimization than as independent selection. TCP-MCP improves over graph-generation baselines and exposes Pareto trade-offs among accuracy, cost, and complexity. The ablations show that prompt--topology co-evolution outperforms the prompt-first topology-search variant.

\noindent{\textbf{Limitations.}}
TCP-MCP is evaluated under a fixed backbone and development subsets, which keeps the comparison controlled but does not cover all model families, task distributions, or search budgets. The selected operating points can require more tokens than lower-cost baselines, although the search exposes this trade-off. Our experiments focus on static topologies selected before evaluation, leaving dynamic communication, tool interaction, and long-horizon adaptation for future study.

\noindent{\textbf{Future Work.}}
Future work may extend this formulation to dynamic interaction policies, tool use, external memory, broader backbone studies, and more systematic budget and initialization analyses. Another direction is online adaptation, where prompt and topology updates are conditioned on task uncertainty, intermediate disagreement, or observed failure modes during deployment.

\bibliographystyle{plainnat}
\bibliography{references/references}

\appendix
\input{appendix}

\end{document}

%% file: appendix.tex
\section{Related Work}
\label{app:relatedworks}

\subsection{Single-Agent Reasoning}

Single-agent reasoning methods improve the inference process of a single LLM without introducing explicit inter-agent communication. Zero-shot prompting is the simplest baseline, where a single LLM directly answers the question without decomposition or collaboration. Complexity-based prompting encourages more detailed intermediate reasoning for multi-step problems~\citep{fu2023complexitybased}, while Self-Consistency samples multiple reasoning paths and selects the most consistent answer~\citep{wang2023selfconsistency}. Progressive-Hint Prompting studies how to improve reasoning within a single-model workflow through iterative hints~\citep{zheng2024progressivehint}. These methods are often strong and cost-effective, but they do not model role specialization, inter-agent communication, or topology-dependent information flow.

\subsection{Multi-Agent Collaboration and Aggregation}

Multi-agent methods improve reasoning by introducing multiple agents, intermediate communication, or output aggregation. Fixed-topology baselines such as Complete Graph and Random Graph, following the setting of G-Designer~\citep{zhang2025gdesigner}, test whether multi-agent communication alone is sufficient to improve performance. Complete Graph allows dense communication among agents, while Random Graph uses sampled communication structures as a topology baseline. Since these structures are not jointly adapted with node-level prompts, they may spend many tokens on unnecessary communication or fail to route useful information to the agents that need it. Interaction and aggregation methods such as Multiagent Debate and LLM-Blender show that proposal, critique, revision, and output aggregation can improve reasoning quality~\citep{du2024multiagentdebate,jiang2023llmblender}. General multi-agent frameworks such as AutoGen, ChatDev, and MetaGPT provide abstractions for roles, message passing, and collaborative execution~\citep{wu2024autogen,qian2024chatdev,hong2024metagpt}. These systems reduce the engineering cost of building multi-agent workflows, but they are not primarily designed to optimize communication topology and node-level prompts as a coupled search problem.

\subsection{Automated Topology and Workflow Design}

Automated topology-design methods are the closest line of work to TCP-MCP because they search or optimize multi-agent communication structures. GPTSwarm formulates language-agent systems as optimizable graphs~\citep{zhuge2024gptswarm}, and G-Designer designs multi-agent communication topologies through graph neural networks~\citep{zhang2025gdesigner}. These methods recognize that topology affects reasoning quality and can outperform manually fixed communication structures in some settings. However, they usually focus on graph structure while treating prompts or roles as fixed, reusable, or separately specified components. TCP-MCP instead represents communication topology and node-level role--prompt assignments as a unified genome. The key distinction is that topology edits, structural crossover, role mutation, prompt regeneration, prompt inheritance, and runtime topology-conditioned prompt construction are defined as coupled operations over the same executable search object. This coupled formulation reflects the fact that an agent's behavior depends not only on its stored prompt template, but also on the upstream messages determined by the topology. Unlike graph-level optimization that treats prompts as node-local parameters and edges as separate orchestration variables, TCP-MCP defines variation operators over executable prompt--topology genomes. This means that structural crossover, role mutation, prompt regeneration, and runtime neighbor-conditioned prompt construction are evaluated as coupled changes to the same system, rather than as independent node or edge updates.

\section{Formalization Notes}
\label{app:formalization}

The notation in this paper defines the search object, objectives, and adaptive-control signals used by TCP-MCP. These equations are meant to specify the algorithm, not to provide theoretical guarantees. We do not claim convergence, approximation bounds, or a guarantee that the evolutionary process will find a globally Pareto-optimal solution. The unified genome $M=(G,P)$ defines the joint search space over communication topology and node-level prompting. The multi-objective fitness vector specifies how accuracy, token cost, and structural complexity are compared during selection. The scalar preference score is used only as an auxiliary signal for initialization-time landscape probing. It does not replace Pareto-based environmental selection. Fitness Distance Correlation (FDC) is also computed only after initialization to calibrate the early search bias. During evolution, TCP-MCP does not re-estimate FDC. It instead uses cross-generational Pareto-front diagnostics, including Hypervolume, spacing, frontier gaps, and regional coverage, to adjust exploration.

\section{Implementation Details}
\label{app:implementation}

This section gives the implementation details of TCP-MCP. We describe how structural crossover, mutation, environmental selection, and adaptive control are implemented in the evolutionary search. The discussion complements the main method section and specifies the search procedure at a level that supports reproduction, without turning the appendix into code-level pseudocode.

\subsection{Initial Population Construction}
\label{app:initial_population}

TCP-MCP initializes each run with a diverse population of prompt--topology genomes. We use Latin Hypercube Sampling (LHS) to cover the main design dimensions of the initial search space, including graph size, edge density, topology pattern, role composition, and prompt style. Each LHS sample is then mapped to an executable genome $M=(G,P)$.

For topology initialization, TCP-MCP samples the number of non-input agents from a bounded range and instantiates a directed acyclic communication graph. The initial graph is drawn from a mixture of structural templates, including chain, tree, star, layered, and sparse random DAG patterns. The sampled LHS coordinate controls edge density. After graph construction, TCP-MCP checks reachability from the input node to the final decision node and rejects edges that violate the acyclic execution order. If a sampled graph is invalid, the implementation either resamples it or repairs it by adding the minimum links needed to make the candidate executable.

For role initialization, each non-input node receives a role from a task-conditioned role pool. The pool contains task-specific expert roles, domain-heuristic roles, and general reasoning roles. For MMLU and MMLU-Pro, the task-specific pool includes Question Decomposer, Option Verifier, Knowledge Checker, Definition Anchor, Trap Detector, Counterfactual Reasoner, and Consistency Checker. For GSM8K, it includes mathematical roles such as Math Solver, Mathematical Analyst, Programming Expert, Inspector, Formula Auditor, and Unit Checker. General roles, including Planner, Critic, Verifier, Summarizer, and Aggregator, are shared across datasets. The final decision or aggregation role is fixed when needed, so that all candidates expose the same external prediction interface.

Prompt initialization follows the assigned role and dataset domain. For each role, TCP-MCP either uses a role-conditioned seed template from the prompt registry or generates an initial instruction with the same PromptMutator used in later role mutation. The generated prompt specifies the role definition, expected reasoning behavior, input usage, and output format. Node-specific topology information is not stored in the prompt template. Instead, it is injected at runtime through $\textsc{BuildPrompt}$ according to the node's incoming neighbors. This keeps the initial prompt representation consistent with the rest of the evolutionary process.

The LHS population is therefore diverse along both structural and semantic dimensions. Candidates differ in graph size, communication pattern, edge density, role assignment, and prompt behavior. After initialization, all candidates are evaluated once on the development subset. Their fitness values are used to compute the initialization-time preference score and Fitness Distance Correlation (FDC), which calibrates the early crossover bias before the main evolutionary loop begins.

\subsection{Structural Crossover and Prompt Minimal Inheritance}

TCP-MCP performs structural crossover by recombining topology-level modules, rather than by mixing prompt strings directly. Given two parent genomes, the operator treats one parent as the recipient and the other as the donor. It selects a subgraph module from the donor and transplants it into the topology of the recipient. The recipient keeps the rest of its original graph, while the donor provides the selected structural module. The resulting topology is checked before acceptance, and any crossover output that violates the directed acyclic communication constraint is rejected. The module-matching procedure first tries to align nodes by exact role match. When exact role matching is not available, the operator uses a structural anchor based on boundary degree, defined as the sum of the in-degree and out-degree of a node. This fallback favors high-interface nodes, which are more likely to preserve useful information flow after module transplantation. This allows crossover to remain applicable even when the two parents do not have identical role assignments.

Prompt inheritance follows Prompt Minimal Inheritance (PMI). Nodes retained from the recipient parent keep their original roles and prompt templates unchanged. Nodes introduced from the donor module inherit the donor-side role and prompt template when a valid donor prompt assignment is available. If a transplanted donor node does not have a valid prompt--role entry, the implementation uses a lightweight role-conditioned default prompt instead of invoking full prompt regeneration. In this way, crossover can make substantial changes to the structural context while changing prompt text only when necessary. This design reflects how prompts are represented in TCP-MCP: prompt templates are stored independently of the concrete topology, and topology-dependent context is added only at runtime through the prompt-construction function. The same inherited prompt template can therefore lead to different behavior after crossover, because the node may receive different upstream messages. PMI preserves useful local prompt behavior while still allowing the surrounding communication structure to change.

\subsection{Mutation Operators}
\label{app:mutation_details}
TCP-MCP uses three classes of mutation: local topology mutation, role mutation with prompt regeneration, and radical structural mutation. Local topology mutation edits the communication graph through elementary operations, including edge addition, edge removal, node addition, and node removal. Each proposed structural edit is checked before acceptance. Edits that introduce cycles are rejected, so the candidate system remains executable under the feed-forward communication protocol. Role mutation changes the functional identity of a node. The candidate role pool is domain-aware and is assembled from dataset-specific expert roles, domain-heuristic roles, and general-purpose reasoning roles. For MMLU and MMLU-Pro, the expert pool includes roles such as Question Decomposer, Option Verifier, Knowledge Checker, Trap Detector, Definition Anchor, Counterfactual Reasoner, and Consistency Checker. For GSM8K, the pool includes mathematical roles such as Math Solver, Mathematical Analyst, Programming Expert, and Inspector. Domain-heuristic and general-purpose roles include Calculator, Mathematician, Statistician, Quantitative Analyst, Debugger, Tester, FactChecker, StepPlanner, FormulaAuditor, UnitChecker, ConflictDetector, ConstraintExplicator, SemanticDifferentiator, and GroundingEnforcer. Output-critical roles, such as final decision or output aggregation roles, are excluded from mutation to avoid changing the external behavior of a candidate system.

When a role changes, TCP-MCP regenerates the corresponding prompt body instead of only replacing the role name. PromptMutator produces a new instruction body conditioned on the new role and the dataset domain. The generated prompt must start with the new role definition, include explicit execution actions, and satisfy role-type-specific requirements: checker roles must include verification behavior, solver roles must include concrete solution steps, and constraint-oriented roles must explicitly enforce task constraints. Node identity is not included in the generated prompt text, because node-specific topology information is added at runtime. Prompt regeneration uses a bounded retry mechanism. If all retries fail, the system first uses a role-specific registry template when one exists, and otherwise uses a minimal role-conditioned fallback prompt. Radical mutation is used when ordinary local edits are not sufficient. It can introduce larger structural changes, such as chain, tree, star, or layered topologies, and its probability is controlled by the adaptive control index described in Section~\ref{app:adaptive_control}. In practice, radical mutation is rare when the search is progressing well and becomes more likely when the Pareto frontier stagnates or the population loses diversity.

\subsection{Environmental Selection and Diversity Preservation}
\label{app:selection_details}
TCP-MCP uses constrained NSGA-II for environmental selection. Each candidate genome is evaluated with a multi-objective fitness vector over task accuracy, token cost, and structural complexity. Accuracy is maximized, whereas token cost and complexity are minimized. Since low-cost candidates are not useful when their accuracy is too low, TCP-MCP applies a dynamic feasibility constraint before standard Pareto ranking. The feasibility threshold is defined relative to the current best accuracy. A candidate is feasible only if its accuracy lies within a fixed margin of the current best candidate. Feasible candidates dominate infeasible candidates regardless of token cost or complexity, preventing the search from selecting very cheap but ineffective systems only because they occupy favorable cost regions. After feasibility filtering, candidates are ranked by standard non-dominated sorting. Within each Pareto front, crowding distance favors candidates in less crowded regions of the objective space. The tie-breaking order is feasibility, Pareto rank, crowding distance, bin capacity, and deterministic iteration order.

TCP-MCP further preserves diversity through bin-based niching, a structural quota, and an external elite archive. Bin-based niching partitions the cost--complexity space using quantile-based boundaries, and each bin can contribute only a limited number of elites. This prevents a dense region of the Pareto frontier from dominating the selected population. The bin cap depends on the diversity deficit: it remains stricter when diversity is low and gradually relaxes when the population is sufficiently diverse. The structural quota prevents premature collapse toward trivial low-node topologies by retaining structurally richer candidates when they satisfy the dynamic accuracy floor and remain competitive under Pareto ranking. The external elite archive can reintroduce validated elites from earlier generations with duplicate checks. If archive injection cannot fill the reserved slots, the implementation falls back to cloning selected elites with duplicate filtering. These mechanisms allow the search to recover useful genomes that may have disappeared from the current generation.

\subsection{Adaptive Control}
\label{app:adaptive_control}

TCP-MCP uses adaptive control to regulate exploration during evolution. The controller does not replace Pareto-based selection. It adjusts mutation pressure and targeted injection behavior using cross-generational diagnostics of the Pareto frontier and the selected population. The main adaptive signal is a scalar control index $s \in [0,1]$, which combines hypervolume progress, elite-best accuracy progress, token-cost pressure, and diversity deficit. Each component is clipped before aggregation, and the final control index is also clipped to $[0,1]$. The weighting scheme changes across generations: early generations give more weight to diversity preservation, while later generations shift toward hypervolume and accuracy. This schedule gives a simple transition from exploration to exploitation. The control index is mapped to the radical-mutation probability through the sigmoid function in Eq.~\ref{eq:stagnation_probability}, where the implementation uses a steep sigmoid. Radical mutation is therefore unlikely under mild pressure, but its probability rises quickly when the control index becomes large.

The controller also adjusts topology and role mutation rates from their base values using the control index and runtime landscape features, including diversity, neutrality, multimodality, and ruggedness proxies computed from recent population history. These runtime features are distinct from the initialization-time Fitness Distance Correlation (FDC). FDC is computed once after initialization and is used only to calibrate early crossover bias; it is not repeatedly estimated during evolution. At the population level, TCP-MCP can perform targeted injection. The targeted injector monitors Pareto-front diagnostics, including hypervolume growth, spacing, maximum frontier gap, and regional coverage. When it detects a recognizable failure mode, it generates repair-oriented genomes: regression recovery introduces conservative candidates based on recently strong regions, multi-island split introduces bridging structures, semantic deficiency targets missing Pareto regions, diversity loss introduces structurally varied candidates, and radical innovation introduces more exploratory reasoning structures. These injected candidates are conditioned on the diagnosed failure mode and the underrepresented Pareto region, but they must still compete under constrained NSGA-II and are retained only if they satisfy the accuracy and Pareto-selection criteria.

\subsection{Prompt Templates and Mutation Examples}
\label{app:prompt_examples}

TCP-MCP stores each node prompt as a role-conditioned template. Topology-dependent context is added only at runtime by \textsc{BuildPrompt}. This section gives representative examples to make the prompt representation and mutation process explicit.

For multiple-choice benchmarks such as MMLU and MMLU-Pro, decomposition and verification roles use short task-specific instructions. A decomposition role classifies the question type, extracts key entities, and states the goal in a fixed JSON format. A verification role independently checks whether each candidate answer is factually correct and relevant to the question, and defers calculation-heavy cases to the math solver. A representative verification template is shown below.

\begin{quote}
\small
\begin{verbatim}
## IDENTITY
You are the Option Fact Checker.

## TASK
1. Fact Check: Check whether each option is factually true.
2. Relevance: Check whether each option answers the question.

## CRITICAL RULE
- If this is a math question, output: "Defer to MathSolver."
- Do not copy other agents' analyses. Check the facts independently.

## OUTPUT
- Option [X]: [Correct/Incorrect] - [Brief Reason]
\end{verbatim}
\end{quote}

For GSM8K, solver prompts are more calculation-oriented. The math solver receives hints from upstream agents, solves the problem step by step, re-reads the final sentence before solving, and must end with a fixed integer-answer format, \texttt{The answer is <integer>}. Few-shot examples, when used, are attached to this template in the released configuration files.

Role mutation regenerates the prompt body instead of only replacing the role name. For example, when a node mutates from a general reasoning role to a trap-detection role, \textsc{PromptMutator} generates an instruction that asks the node to check for unit errors, scope errors, and option-level inconsistencies. This keeps the mutated role aligned with its prompt behavior.

PMI follows the same template-level representation. During crossover, nodes retained from the recipient keep their prompt templates, while transplanted donor nodes inherit the donor-side role and prompt template when available. The stored template contains no edge information or upstream-node identifiers; those are injected by \textsc{BuildPrompt} under the current topology. Generated prompts are subject to hard constraints, including a 3000-character limit, removal of greetings and generic assistant phrases, and the requirement that each prompt specify at least one explicit reasoning behavior, such as elimination, step-by-step solving, or consistency checking.

\section{Sensitivity Analysis}
\label{app:sensitivity}

We include a limited sensitivity analysis to test whether TCP-MCP relies on a narrow set of implementation constants. Full sweeps over evolutionary hyperparameters are costly, so we use the development subsets to examine robustness in search behavior rather than to rank hyperparameter settings. We focus on the main factors that affect exploration: the adaptive-control schedule, population and elite sizes, generation budget, and the complexity objective.

The adaptive controller determines when the search moves from local refinement to stronger exploration. The control index $s$ is mapped to the probability of radical mutation through a sigmoid function. If this mapping is too flat, radical mutation reacts weakly to frontier stagnation; if it is too sharp, small changes near the transition region can make mutation behavior unstable. The setting used in the main experiments keeps radical mutation rare during normal progress and increases it only under persistent stagnation or diversity loss. The weighting schedule follows the same logic. Early generations emphasize diversity to keep multiple topology and cost regimes active, while later generations emphasize hypervolume and accuracy to refine the Pareto frontier.

Population size, elite size, and generation budget control the cost--coverage trade-off of evolutionary search. Smaller populations and elite sets reduce evaluation cost but can discard useful structures before they are recombined or refined. Larger settings improve coverage but increase token use directly and leave less room for efficient search under a fixed budget. In the main configuration, bin-based niching and the structural quota reduce this sensitivity by preventing one cost--complexity region from dominating the elite set and by avoiding early collapse to trivial low-node graphs. Across the monitored runs, most hypervolume gains occur in the early and middle generations, after which the frontier tends to stabilize. This pattern suggests that the chosen generation budget is sufficient to reach a stable trade-off region under the available evaluation budget.

The complexity objective controls pressure against unnecessarily large communication graphs. If this pressure is too weak, the search may retain systems whose additional nodes and edges bring little accuracy gain. If it is too strong, the population may collapse to small topologies before useful collaborative structures are explored. We therefore treat structural complexity as a Pareto objective rather than a hard graph-size constraint. The structural quota only guards against premature collapse: structurally richer candidates must still satisfy the dynamic accuracy floor and remain competitive under Pareto ranking. Final operating points are selected from the development Pareto frontier to reflect the accuracy--cost--complexity trade-off, not to favor the largest or most expensive candidate found during search.

\section{Compute Resources}
\label{app:compute_resources}

TCP-MCP does not require local model training or GPU-based fine-tuning. All experiments can be launched from a local machine or a standard CPU worker, with the main computational cost coming from external LLM API inference. We therefore report token usage, API-call counts, and wall-clock time as the primary compute-cost measures. The token counts reported in the main result table correspond to the final held-out evaluation of the selected Pareto operating point, not to the full evolutionary search cost. The evolutionary search itself is performed on the fixed development subsets and incurs additional API inference cost when candidate genomes are evaluated.

\begin{table}[t]
\centering
\caption{Compute summary for the selected TCP-MCP systems on held-out evaluation. Token usage corresponds to the final held-out evaluation reported in Table~\ref{tab:main_results}. API calls are approximate and are estimated as the number of held-out examples multiplied by the number of LLM-executed nodes in the selected topology.}
\label{tab:compute_resources}
\begin{tabular}{lrrrr}
\toprule
Dataset & Held-out examples & Selected nodes & Tokens & API calls \\
\midrule
MMLU & 13{,}961 & 3 & 88.03M & 41{,}883 \\
MMLU-Pro & 12{,}032 & 3 & 97.28M & 36{,}096 \\
GSM8K & 1{,}240 & 2 & 6.02M & 2{,}480 \\
\bottomrule
\end{tabular}
\end{table}

The API-call counts in Table~\ref{tab:compute_resources} are for final held-out evaluation only. We count one LLM call for each executed node on each held-out test example. These counts exclude failed requests, retries, preliminary experiments, baseline reruns, and all API calls used during evolutionary search on the development subsets. Search-time cost depends on the number of evaluated genomes, the active nodes in each candidate topology, the prompt length of each node, and the generation budget. In our implementation, search-time token usage and API calls are recorded in the checkpoint logs and are treated separately from the held-out evaluation cost used for the main accuracy--token comparisons.

To make the search-time inference cost explicit, Table~\ref{tab:search_time_usage} reports token usage and API calls recorded in the main checkpoint files for each benchmark. These logged counts measure LLM inference used for candidate evaluation on the development subset during evolutionary search. Structural crossover does not require LLM inference. Mutation may invoke the LLM for prompt regeneration in some cases, and such calls are tracked separately from the candidate-evaluation calls reported in Table~\ref{tab:search_time_usage}.

\begin{table}[t]
\centering
\caption{Search-time LLM inference usage recorded from the main checkpoint files. The counts correspond to development-set candidate evaluation during evolution and exclude validation folders, elite folders, preliminary experiments, baseline reruns, and final held-out evaluation.}
\label{tab:search_time_usage}
\begin{tabular}{lrr}
\toprule
Benchmark & Logged search-time tokens & Logged search-time API calls \\
\midrule
MMLU & 351.05M & 135{,}880 \\
MMLU-Pro & 486.05M & 174{,}500 \\
GSM8K & 128.02M & 66{,}120 \\
\midrule
Total & 965.12M & 376{,}500 \\
\bottomrule
\end{tabular}
\end{table}

Wall-clock time is reported as an operational measure because it depends on API latency, rate limits, retry behavior, and endpoint availability. Table~\ref{tab:search_time_wallclock} reports the recorded wall-clock time for evolutionary search on each benchmark under our API setting.

\begin{table}[t]
\centering
\caption{Recorded search-time wall-clock cost under our API setting. The times correspond to evolutionary search on the fixed development subsets and exclude final held-out evaluation, preliminary experiments, and baseline reruns.}
\label{tab:search_time_wallclock}
\begin{tabular}{lr}
\toprule
Benchmark & Recorded wall-clock time \\
\midrule
MMLU & 20.0h \\
MMLU-Pro & 25.7h \\
GSM8K & 9.7h \\
\midrule
Total & 55.4h \\
\bottomrule
\end{tabular}
\end{table}

The supplemental code package includes the scripts, configuration files, and evaluation commands used to reproduce the API-based experiments.

\paragraph{Scope and cost.}
TCP-MCP is evaluated under a fixed backbone and fixed development-subset protocol, with final results reported on held-out test sets. This design avoids test-set leakage and keeps comparisons controlled, but broader validation across additional backbones and search budgets remains future work. Since TCP-MCP is a Pareto-based optimizer, the selected high-performance operating points may require more tokens than lower-cost baselines; however, searched prompt--topology patterns can be reused as fixed templates or transferred to related settings, which can amortize part of the initial search cost.

\paragraph{Additional Cost Visualization}
\label{app:acc-cost-visualization}

\begin{figure*}[t]
    \centering
    \includegraphics[width=\textwidth]{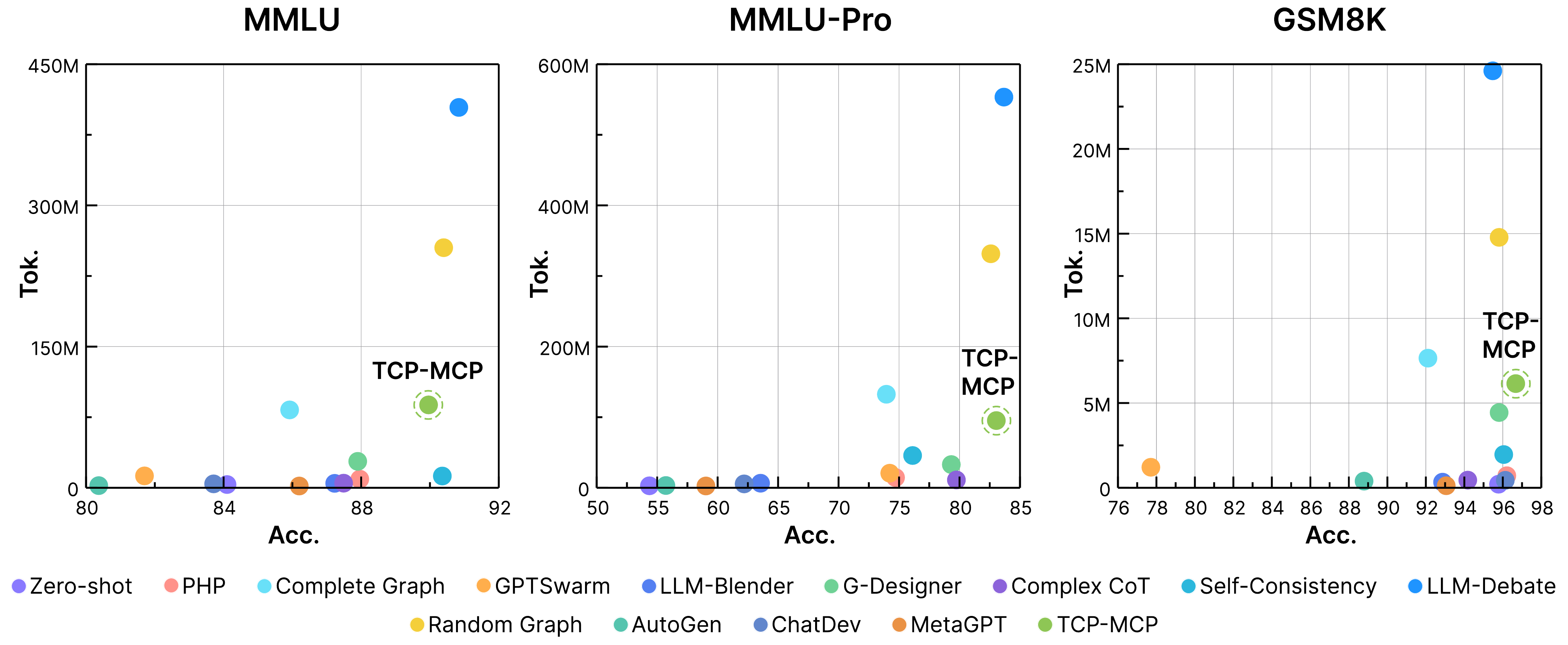}
    \caption{
    Supplementary visualization of accuracy and token cost for the main results in Table~\ref{tab:main_results}.
    Each point corresponds to one method on the held-out evaluation set.
    The horizontal axis shows accuracy, and the vertical axis shows total held-out inference tokens.
    TCP-MCP is highlighted with a dashed circle.
    }
    \label{fig:app_acc_token_tradeoff}
\end{figure*}

Figure~\ref{fig:app_acc_token_tradeoff} visualizes the accuracy and token cost reported in Table~\ref{tab:main_results}.
Across MMLU, MMLU-Pro, and GSM8K, TCP-MCP is not always the lowest-token method, but it remains in a high-accuracy region while avoiding the much larger token costs of high-interaction baselines such as LLM-Debate and Random Graph.
This figure complements the table by showing that the selected TCP-MCP operating points emphasize accuracy while keeping token usage below the most expensive multi-agent protocols.

\section{Second Evaluation on Main Datasets}
\label{app:second_eval}

We conduct a second evaluation on the main datasets as an additional stability check.
This second evaluation repeats inference for the selected systems under the same held-out evaluation protocol, but does not rerun evolutionary search.
As shown in Table~\ref{tab:second_eval}, the second-run results follow the same overall trend as the main results, with TCP-MCP remaining in the high-accuracy region on all three datasets.
We therefore use this check as additional robustness evidence rather than as a multi-seed evolutionary result.

\begin{table}[t]
\centering
\caption{Second held-out evaluation of the selected TCP-MCP systems. The second run repeats inference under the same held-out protocol without rerunning evolutionary search.}
\label{tab:second_eval}
\begin{tabular}{lccc}
\toprule
Dataset & Main result & Second run & Difference \\
\midrule
MMLU & 89.96 & 89.10 & -0.86 \\
MMLU-Pro & 82.66 & 82.33 & -0.33 \\
GSM8K & 96.61 & 96.40 & -0.21 \\
\bottomrule
\end{tabular}
\end{table}

\section{Statistical Uncertainty of Held-Out Accuracy}
\label{app:statistical_uncertainty}

For the main held-out evaluation, we report accuracy on test sets obtained by removing the fixed development subsets from the original benchmark collections. To quantify finite-sample uncertainty in the reported held-out accuracy values, we compute approximate binomial confidence intervals:
\begin{equation}
\hat{p} \pm 1.96
\sqrt{\frac{\hat{p}(1-\hat{p})}{n}},
\label{eq:binomial_confidence_interval}
\end{equation}
where $\hat{p}$ is the observed accuracy and $n$ is the number of held-out test examples. These intervals estimate uncertainty due to finite test-set sampling only. They do not capture variability from evolutionary initialization, LLM stochasticity, API-level nondeterminism, or development-subset selection.

Because full multi-seed evolutionary runs are computationally expensive, our ablation and transfer analyses are interpreted as directional evidence rather than as full multi-seed significance tests. In particular, we use them to check whether removing major components or separating prompt and topology search changes the selected operating points under the same controlled protocol, not to estimate run-to-run variance.

\section{Operating Point Selection}
\label{app:operating_point_selection}

TCP-MCP returns a Pareto set rather than a single system. We therefore select the reported system from the development-set Pareto front before final held-out evaluation. The selection rule is fixed before inspecting held-out results.

Let $\mathcal{P}_{\mathrm{dev}}$ denote the set of non-dominated candidates on the development-set Pareto front. For a candidate $M$, let $A_{\mathrm{dev}}(M)$ be its development accuracy, $C_{\mathrm{dev}}(M)$ be its development token cost, and $K(M)$ be its structural complexity. We first restrict the candidate pool to candidates that satisfy the same dynamic accuracy floor used during environmental selection:
\begin{equation}
\mathcal{P}_{\tau}
=
\left\{
M \in \mathcal{P}_{\mathrm{dev}}
\mid
A_{\mathrm{dev}}(M) \geq A_{\max} - \delta
\right\},
\end{equation}
where $A_{\max} = \max_{M \in \mathcal{P}_{\mathrm{dev}}} A_{\mathrm{dev}}(M)$ and $\delta$ is the accuracy tolerance used in environmental selection.

We then remove high-cost tail candidates unless their additional token cost gives a meaningful development-accuracy gain. Specifically, let $q_{0.8}$ be the 80th percentile of token cost among candidates in $\mathcal{P}_{\tau}$. A candidate $M \in \mathcal{P}_{\tau}$ is removed if it satisfies both conditions:
\begin{equation}
C_{\mathrm{dev}}(M) \geq q_{0.8},
\end{equation}
and
\begin{equation}
\max_{M' \in \mathcal{L}(M)}
A_{\mathrm{dev}}(M')
\geq
A_{\mathrm{dev}}(M) - \epsilon_{\mathrm{acc}},
\end{equation}
where
\begin{equation}
\mathcal{L}(M)
=
\left\{
M' \in \mathcal{P}_{\tau}
\mid
C_{\mathrm{dev}}(M') \leq 0.8 C_{\mathrm{dev}}(M)
\right\}.
\end{equation}
We set $\epsilon_{\mathrm{acc}} = 0.015$, corresponding to 1.5 percentage points. Thus, a high-cost candidate is kept only when it improves development accuracy by more than 1.5 percentage points over candidates that use at least 20\% fewer tokens.

Let the remaining candidate set be $\mathcal{P}_{\mathrm{sel}}$. The final operating point is selected as:
\begin{equation}
M^{\star}
=
\arg\max_{M \in \mathcal{P}_{\mathrm{sel}}}
\left(
A_{\mathrm{dev}}(M),
- C_{\mathrm{dev}}(M),
- K(M)
\right),
\end{equation}
where the tuple is compared lexicographically after the high-cost-tail filtering above. In other words, among the remaining candidates, we first select the candidate with the highest development accuracy, then use lower token cost and lower structural complexity as tie-breakers. This rule is applied only on the development set.

This protocol reflects the multi-objective nature of TCP-MCP. The selected system is not necessarily the cheapest point or the highest-token point on the Pareto front. Instead, it is a high-accuracy operating point that avoids spending many additional tokens for marginal gains on the development set. The held-out test set is used only after this selection step, so test performance does not affect the selected genome.
\section{Ablation Study Details}
\label{app:ablation_details}

We use ablations to separate the effects of the main search components in TCP-MCP. Each variant changes one part of the evolutionary procedure while keeping the backbone model, evaluation protocol, development subset, and held-out test setting the same as in the main experiments. Unless stated otherwise, all variants use the same population size, elite size, Pareto-based selection rule, and final development-set operating-point selection rule as full TCP-MCP.

AB1 removes adaptive rate control. The crossover rate, topology mutation rate, and role mutation rate are kept fixed throughout evolution, so the search no longer adjusts these rates using stagnation or diversity signals after initialization. FDC is still computed at initialization, but it is not used for cross-generational control. AB2 removes structural crossover. Offspring are generated only through mutation, which prevents the search from recombining topology subgraphs with their associated roles and prompts. This disables the main mechanism for exchanging co-adapted prompt--topology modules. AB3 removes mutation-related operations. The search then relies only on structures inherited through crossover, without topology edits, role changes, or prompt regeneration. This limits both local refinement and broader exploration after the initial population is constructed.

AB4 replaces simultaneous co-evolution with sequential prompt-first topology search. It uses the same population size, elite size, development subset, Pareto-based selection rule, and total evolutionary budget as the full method. In AB4, topology is fixed while prompts are optimized for 11 generations, and the selected development-set Pareto solution from this stage provides the fixed prompt configuration for a second 10-generation topology-optimization stage. The final system is selected from the development-set Pareto front using the same operating-point rule as full TCP-MCP, and is then evaluated on the held-out test set. Thus, AB4 optimizes both design spaces under the same overall budget, but does not allow prompts and topology to change jointly within a single evolutionary phase.

These variants test different parts of the search process. AB1 examines whether adaptive control improves over a fixed evolutionary schedule. AB2 tests whether module-level structural recombination is important for prompt--topology co-adaptation. AB3 tests whether mutation is needed for continued exploration and refinement. AB4 tests whether the gains of TCP-MCP require simultaneous prompt--topology co-evolution, or whether they can be recovered by optimizing prompts first and topology second. Together, these ablations evaluate the main design claim of TCP-MCP: prompts and communication topology should be optimized as a coupled system rather than as independent or fixed components.

Removing structural crossover causes the largest degradation on MMLU-Pro and a clear drop on MMLU, suggesting that module-level recombination is important for combining useful prompt--topology modules. On MMLU-Pro, AB2 lowers accuracy by 2.51 percentage points relative to full TCP-MCP and substantially increases token usage. This result suggests that structural crossover is not only an efficiency mechanism, but also a key way to combine useful prompt--topology modules. AB3 also weakens the search, indicating that mutation is needed to introduce communication structures, role assignments, and prompt variants that cannot be obtained from recombination alone.

The sequential variant clarifies the role of joint optimization. AB4 still optimizes both prompts and topology, but it does so in separate stages. Its performance therefore tests whether a prompt-first two-stage procedure can approximate the coupled search used by TCP-MCP. When this variant underperforms the full system, the gap suggests that useful topology changes depend on the behavior induced by the current prompts, and that optimizing one side while holding the other fixed can lead the search toward suboptimal intermediate designs.

Ablation variants can also shift the selected Pareto operating point and its token profile. For example, AB1 sometimes uses fewer tokens than the full system, but this does not imply a better trade-off between accuracy and cost because it also reduces accuracy. The stronger degradations under variants that remove operators or use prompt-first sequential optimization provide the main evidence: removing structural recombination, mutation, or simultaneous prompt and topology updates weakens the ability of TCP-MCP to find effective prompt and topology systems.

\section{Hypervolume Diagnostics}
\label{app:hv_diagnostics}

Table~\ref{tab:hv_diagnostics} reports normalized Hypervolume (HV) at generations 1, 11, and 21. These checkpoints summarize the Pareto-front trend after the first generation, in the middle of evolution, and at the final reported generation.

\begin{table}[t]
\centering
\small
\caption{Normalized Hypervolume (HV) diagnostics across generations.}
\label{tab:hv_diagnostics}
\begin{tabular}{lccc}
\toprule
\textbf{Dataset} & \textbf{Gen. 1} & \textbf{Gen. 11} & \textbf{Gen. 21} \\
\midrule
MMLU     & 0.5905 & 0.6945 & 0.6963 \\
MMLU-Pro & 0.3698 & 0.5036 & 0.6098 \\
GSM8K    & 0.7939 & 0.8399 & 0.8431 \\
\bottomrule
\end{tabular}
\end{table}